\documentclass[runningheads]{llncs}

\usepackage{graphicx}
\usepackage{caption}
\usepackage{xspace}

\usepackage{graphicx}
\usepackage{graphicx}
\usepackage{amssymb}
\usepackage{xcolor}
\usepackage{pifont}
\usepackage{enumerate}
\usepackage{amsmath}
\usepackage{booktabs}
\usepackage{multirow}
\usepackage{subfigure}
\usepackage{ulem}
\usepackage[misc]{ifsym}
\newcommand{\sys}{\textsc{TreeCSS}\xspace}
\newcommand{\starall}{\textsc{StarALL}\xspace}
\newcommand{\treeall}{\textsc{TreeALL}\xspace}
\newcommand{\starcss}{\textsc{StarCSS}\xspace}

\makeatletter
\def\thanks#1{\protected@xdef\@thanks{\@thanks
        \protect\footnotetext{#1}}}
\makeatother

\begin{document}

\newcommand{\stitle}[1]{\vspace*{0.4em}\noindent{\bf #1\/}}
\title{\sys: An Efficient Framework for Vertical Federated Learning}
% \author{Anonymous submission}
\author{Qinbo Zhang\inst{1}* \and
        Xiao Yan\inst{2}*   \and
        Yukai Ding\inst{1} \and
        Quanqing Xu\inst{3} \and
        Chuang Hu\inst{1\textrm{\Letter}} \and
        Xiaokai Zhou\inst{1} \and
        Jiawei Jiang\inst{1\textrm{\Letter}} 
        \thanks{* Equal contribution \textrm{\Letter} Corresponding author}}
%
% \authorrunning{Anonymous Submission}
\institute{School of Computer Science, Wuhan University \\ 
\email{\{qinbo\_zhang, yukai.ding, handc, xiaokaizhou, jiawei.jiang\}@whu.edu.cn}  
\and
Centre for Perceptual and Interactive Intelligence (CPII)\\
\email{yanxiaosunny@gmail.com}  
\and
OceanBase  \\
\email{xuquanqing.xqq@oceanbase.com}}

\maketitle              % typeset the header of the contribution
\begin{abstract}

Vertical federated learning (VFL) considers the case that the features of data samples are partitioned over different participants.
VFL consists of two main steps, i.e., identify the common data samples for all participants (\textit{alignment}) and train model using the aligned data samples (\textit{training}). However, when there are many participants and data samples, both alignment and training become slow. As such, we propose \sys{} as an efficient VFL framework that accelerates the two main steps. 
In particular, for sample alignment, we design an efficient multi-party private set intersection (MPSI) protocol called \textit{Tree-MPSI}, which adopts a tree-based structure and a data-volume-aware scheduling strategy to parallelize alignment among the participants.
As model training time scales with the number of data samples, we conduct \textit{coreset selection} (CSS) to choose some representative data samples for training. Our CCS method adopts a clustering-based scheme for security and generality, which first clusters the features locally on each participant and then merges the local clustering results to select representative samples. 
In addition, we weight the samples according to their distances to the centroids to reflect their importance to model training.
We evaluate the effectiveness and efficiency of our \sys framework on various datasets and models.
The results show that compared with vanilla VFL, \sys accelerates training by up to 2.93$\times$ and achieves comparable model accuracy.

\end{abstract}
\section{Introduction}
Federated learning (FL)~\cite{konevcny2016federated,li2020federated,fu2022towards} is a paradigm of distributed machine learning that trains models over multiple participants without sharing the data samples. FL encompasses two main scenarios, i.e., horizontal FL (HFL)~\cite{zhang2021survey,jiang2022distributed,miao2021heterogeneity} and vertical FL (VFL)~\cite{yang2019federated,liu2019communication,fu2021vf2boost,jiang2022vf}. In HFL, the participants hold different data samples; while in VFL, each participant has a separate set of features for all data samples. We focus on VFL in this paper, which usually emerges when different institutions possess different data for common entities and has applications in areas such as healthcare, finance, and IoT. For instance, each bank holds its own financial records of the customers, and the banks can collaborate via VFL to train a model to predict the financial risk of the customers~\cite{mammen2021federated}.

VFL involves two main steps, i.e., \textit{data alignment} and \textit{model training}. With data alignment, the participants agree on a common set of data samples to use for training. This is necessary because the participants may have non-overlapping samples or assign different identifiers for the samples; and the alignment step derives a unique global identifier for each sample. To protect data privacy, VFL usually adopts \textit{private set intersection} (PSI) methods for data alignment~\cite{hazay2017scalable}. For model training, VFL needs to exchange activations and gradients for each sample (called \textit{instance-wise communication})~\cite{ceballos2020splitnn}. This is because the features are partitioned over the participants, and thus each participant can only compute part of the activations and gradients for a sample.

\noindent \textbf{Challenges.}
Both data alignment and model training become slow when there are many participants and data samples. 
This is because the computation and communication costs of PSI grow quadratically with the number of data samples and linearly with the number of participants. For model training, VFL needs to communicate activations and gradients for each data sample as discussed above, and thus the communication costs scale linearly with the number of training samples. An efficient VFL framework should tackle these scalability challenges.

\noindent \textbf{Current research landscape.} 
For data alignment, VFL usually utilizes multi-party private set intersection (MPSI)~\cite{miyaji2015scalable,inbar2018efficient,bay2021practical}. However, the efficiency of existing MPSI approaches suffers when there are many participants with possibly unbalanced data volumes. This is because these approaches mainly use path and star topological structures to coordinate the participants, which hinder parallel computation and cannot adjust the roles of the participants based on their data volumes. \ding{182} In particular, with a path-like structure, each participant conducts sequential two-party PSI with its adjacent party, requiring $O(m)$ rounds with $m$ participants. Due to the lack of parallel computation, the efficiency is poor. \ding{183} The star structure employs a central participant that interacts with all the other participants. It needs only $O(1)$ round but requires high communication bandwidth and computation power for the central participant, which may become the bottleneck and a major failure risk.

Model training can be accelerated by reducing the number of training samples. For such purpose, existing works~\cite{bachem2017practical,feldman2020turning} propose the concept of \textit{coreset}, which selects some representatives from data samples to conduct training without degrading the accuracy of the trained models. V-coreset~\cite{huang2022coresets} adapts coreset for VFL but suffers from two critical limitations. 
\ding{182} V-coreset cannot ensure data privacy because it directly sends the original labels to the server for model training, potentially resulting in label leakage for the samples.
\ding{183} V-coreset is not general as it tailors construction to specific machine learning models and only supports linear regression and unsupervised clustering, excluding significant classification models.

Given the limitations of existing works, we ask the following research question. \textit{Is it possible to design an efficient VFL framework that accelerates the two key steps, ensures data privacy, and generalizes across machine learning models?}

\begin{figure*}[!t]
\centering
\includegraphics[width=0.75\textwidth]{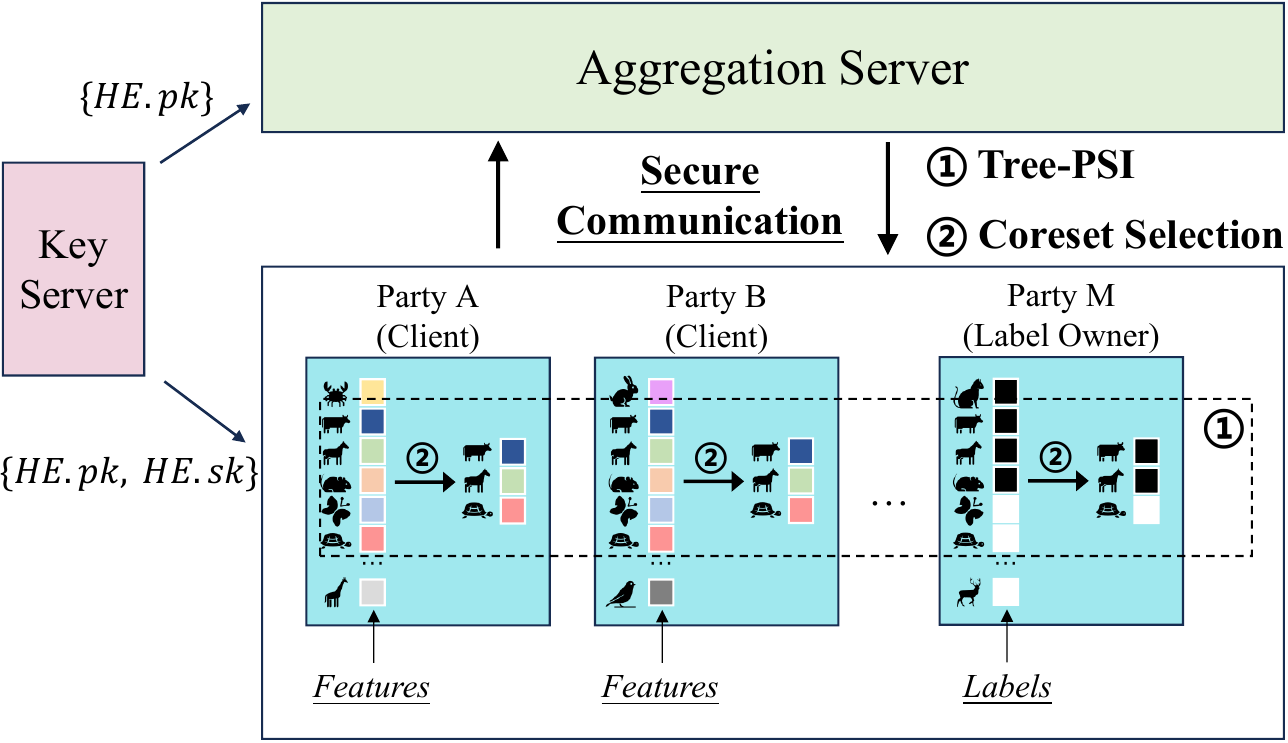}
\caption{An illustration of our \sys{} framework.} 
\label{fig:treecss}
\end{figure*}

\noindent \textbf {Our solution \sys.} By tackling the research question above, we propose \sys{} in Figure~\ref{fig:treecss} as a general and efficient VFL framework that accelerates both PSI and coreset construction. 
\sys addresses three technical challenges.
\ding{182} How to align the data samples efficiently across many participants before constructing the coreset?   
\ding{183} How to choose the coreset samples efficiently? 
\ding{184} Samples in the coreset may contribute differently to the knowledge of model learning, how to acount for the contributions of different samples?
\sys addresses these challenges by using a \textit{tree-based multi-party PSI protocol} (\textit{Tree}-MPSI) for data alignment, a \textit{clustering-based method} (\textit{Cluster}-Coreset) for multi-party coreset construction, and a weighting strategy to consider the importance of the coreset samples for model training.
In particular, \textit{Tree}-MPSI employs a tree-based structure to pair participants for conducting two-party PSI computations and optimizes scheduling based on the data volume of each participant. As coreset selects representative samples, our idea is to first organize the samples into groups with each group containing samples similar to each other, and then choose one sample to represent each group. Our Cluster-Coreset uses $K$-Means, a simple yet effective clustering algorithm, to group the features on each participant, and measure the global similarity of the samples according to their local clustering results. After that, we reweight the samples in the coreset by considering their distances to the cluster centroids across all clients.

To summarize, our contributions include the follows:

\begin{itemize}

\item We observe that the two main steps of vertical federated learning, i.e., data alignment and model training, suffer from efficiency problems when there are many participants and data samples.

\item We propose \sys as a complete vertical federated learning framework that accelerates the two main steps. In particular, Tree-MPSI improves efficiency by scheduling the participants intelligently, and Cluster-Coreset adopts K-means to achieve model generality and data security.

\item We evaluate our \sys on six datasets with diverse proprieties and experiment with both classification and regression models. 
The results show that \sys consistently accelerates vanilla VFL by a large margin and achieves similar model accuracy.

\end{itemize}

\section{Related Work}

\stitle{Vertical federated learning (VFL).} 
VFL trains models on samples whose features are partitioned among the clients, and many works propose models and algorithms for VFL. In particular, Hardy~\cite{hardy2017private} introduces a VFL framework that uses homomorphic encryption to train logistic regression (LR) models. 
Yang~\cite{yang2019quasi} extends this framework by adopting the quasi-Newton method to reduce communication costs. 
Inspired by split learning~\cite{vepakomma2018split,ceballos2020splitnn}, SplitNN divides models into multiple parts to support complex networks in VFL. We speedup VFL under the SplitNN framework with efficient data alignment and coreset construction. 

\stitle{Private set intersection (PSI).} Sample alignment protocols are crucial for VFL because the clients may have non-overlapping samples. PSI protocols are widely used in secure multi-party computation and enable the parties to determine the intersection of their sets without revealing their items.
Two-party PSI protocols are extensively studied and can be realized using techniques such as oblivious transfer \cite{pinkas2014faster}, garbled circuits \cite{huang2012private}, classical public-key cryptosystems \cite{de2009practical}, etc. Some works extend two-party PSI to the multi-party.

In the current landscape, the topological structures of Multi-party Private Set Intersection (MPSI) are centered around \textit{path} and \textit{star} configurations. Hazay~\cite{hazay2017scalable} proposes a protocol that uses the polynomial set encoding and star topology. They evaluate the polynomials obliviously using an additive homomorphic threshold cryptosystem, and provide an extension of the protocol secure in the malicious model. 
Kavousi~\cite{kavousi2021efficient} constructs a protocol including oblivious transfer (OT) extension and garbled Bloom filter as its main ingredients. The protocol involves two interaction types: one uses a star-like communication graph, where one party acts as the sender interacting with all others via OTs. The other utilizes a path-like communication graph, passing a garbled Bloom filter from the first to the last neighboring party. However, only the sender is able to obtain the final PSI result. Vos~\cite{vos2022fast} designs private AND operation among multiple participants by utilizing elliptic curve cryptography, which makes it suitable for MPSI. The participants communicate strictly in a star topology.

\stitle{Coreset.} Coreset selects some representatives from the samples while ensuring that models trained on the coreset yield similar accuracy as full-data training. As such, coreset can reduce the number of training samples and is used to accelerate  machine learning models including clustering \cite{feldman2011unified}, regression \cite{drineas2006sampling}, low-rank approximation \cite{cohen2017input}, and mixture modeling \cite{lucic2017training}. V-coreset extends coreset to VFL~\cite{huang2022coresets} and designs specialized coreset construction methods for linear regression and k-means clustering. In particular, for linear regression, V-coreset computes the orthonormal basis for the features on each client and learns the model according to the projections of the samples on these basis. However, by exchanging the projections, V-coreset may leak private data. For k-means, V-coreset computes the local sensitivity to select the samples to include in the coreset.

\section{Preliminaries}
\label{sec:pre}

In this part, we introduce the SplitNN framework, which provides general settings and procedures for VFL.

There are $M$ participants (also called clients), and one client owns the labels, which is called the label owner and denoted as $C_{lo}$. There are also an aggregation server and a key server. To facilitate secure client-server communication, the key server generates public and private homomorphic encryption (HE) keys to encrypt the messages.
The goal is to train a machine learning model on $N$ data samples: $D = \{(x_{i}\in\mathbb{R}^{d}, y_{i}\in\mathbb{N})\}_{i=1}^{N}$, where $x_{i}$ represents the features of the $i\textsuperscript{th}$ sample, and $y_{i}$ is the label. 
We use $\left [ N \right ]=\left \{ 1,\cdots,N \right \} $ to denote the indices  of the samples and $[M] = \{1,\cdots,M\}$ to denote the set of all clients. For classification and regression tasks, the loss function is usually expressed as 
\begin{equation}
    L(D, \theta):=\sum_{i \in[N]}L(f_{{global}}(x_{i}:\theta), y_{i}).
\end{equation}
The goal is to minimize the sample-wise loss function $L(\cdot , \cdot)$ using model $f_{{global}}(x_{i}:\theta)$ parameterized by $\theta$.

Every feature vector $x_i$ is partitioned over the $M$ clients, and a client $m$ holds some features of all the samples, i.e.,  $D_{m}=\{x_{i}^{m} \in \mathbb{R}^{d_m}: m \in [M]\}_{i=1}^{N}$, where $d_m$ denotes the number of local features on client $m$. Thus, we have $\sum_{m=1}^{M} d_m = d$.
All the labels $D_{label} = \{y_{i} \in \mathbb{N}\}_{i=1}^{N}$ are kept on the label owner.
The model $f_{global}(\theta)$ is partitioned into a bottom model $f_{b}(\theta_{b})$ and a top model $f_{t}(\theta_{t})$. Each client possesses a segment of the bottom model $\{f_{b}^{m}(\theta_{b}^{m}): m \in [M]\}$ that works on its local features, while the top model merges the outputs of the local bottom models for final output and is kept on the aggregation server. The procedure of model training works as follows.

\begin{enumerate}
\item[\ding{182}] The local feature vectors (i.e., $\{x_{i}^{m}\}_{i=1}^{N}$) are processed by the clients using their bottom models (i.e., $f_{b}^{m}(\theta_{b}^{m})$) to produce intermediate outputs.

\item[\ding{183}] The aggregation server merges the intermediate outputs, processes them with the top model $f_{t}(\theta_{t})$, and forwards the results to the label owner.

\item[\ding{184}] The label owner computes loss according to the final outputs and the labels, which is used to derive gradients $g_{t}$ for the top model.

\item[\ding{185}]  The aggregation server updates the top model with gradient $g_{t}$ and computes gradients $\{g_{b}^{m}: m \in [M]\}$ for the local bottom models of the clients, which are used by the clients to update their local models.

\end{enumerate}

\section{The \sys Framework}

In this part, we describe our \sys framework. 
The lifecycle of \sys takes three steps, i.e., \ding{182} data alignment with Tree-MPSI, \ding{183} coreset contruction with Cluster-Coreset on the algined samples, and \ding{184} model training on 
the coreset. 
Step \ding{184} follows the SplitNN framework discussed in Section~\ref{sec:pre}, and we focus on \ding{182} and \ding{183}.

\subsection{Tree-MPSI for Data Alignment}
\begin{figure*}
\centering
    \includegraphics[width=0.75\textwidth]{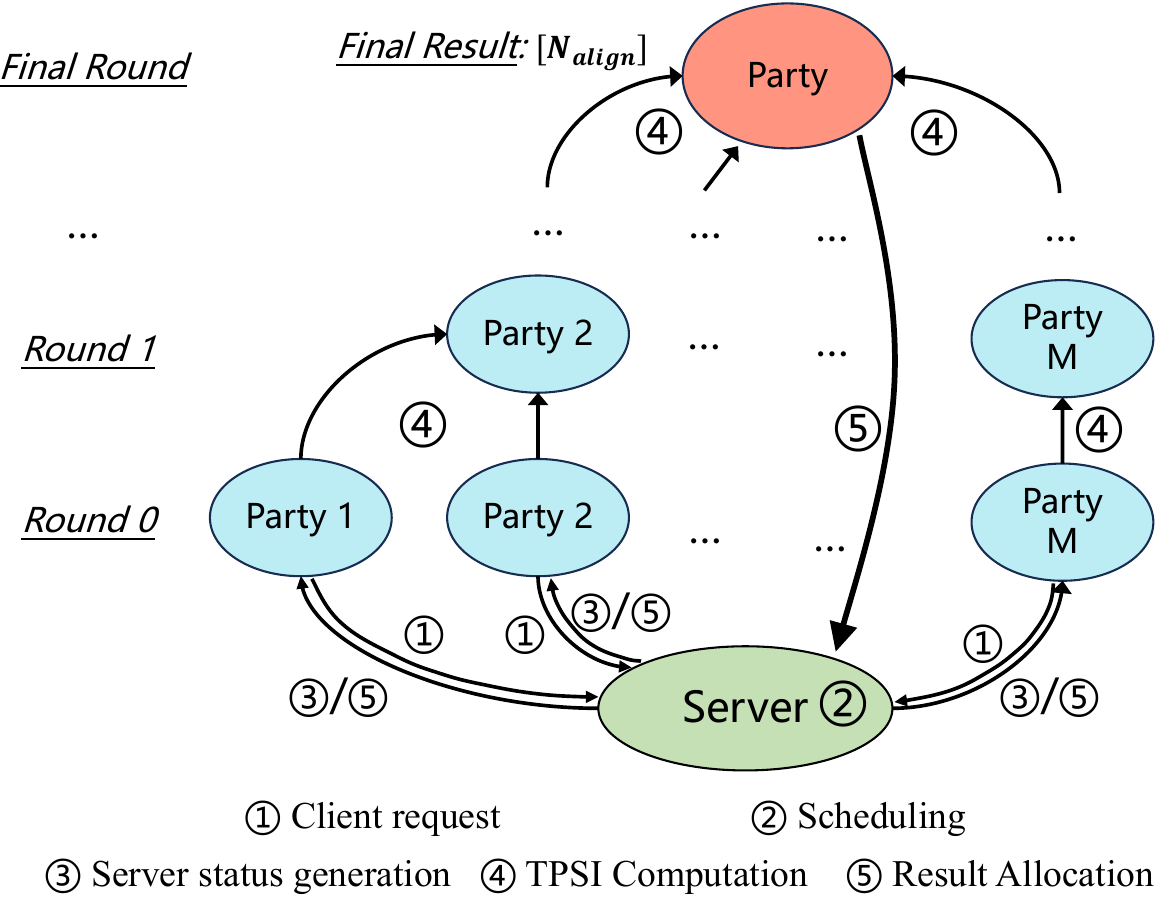}
    \caption{An illustration of Tree-MPSI for data alginement.}
    \label{btmpsi}

\end{figure*}
\noindent  For VFL, data alignment among the clients is essential because their sample orderings may be different. 
For such purpose, we employ private set intersection (PSI) to synchronize the data indices. 
Our Tree-MPSI supports multiple clients and handles many samples efficiently. Figure \ref{btmpsi} provides an illustration of Tree-MPSI. In particular, Tree-MPSI utilizes a tree-based architecture to organize the clients into pairs and schedule two-party PSI. 
For the selection of two-party PSI (TPSI) protocols, we opt for two representative methods: one based on RSA blind signatures and another based on Oblivious Transfer (OT).

\stitle{Two-party PSI primitive.} 
Following, we describe our proposed method using RSA, and the counterpart using OT is similar.
For RSA blind signature method, the protocol involves a \textit{sender} and a \textit{receiver}. The process starts with the sender generating encryption keys and sharing the public key. 
The receiver encrypts its sample indicators using the public key and random numbers, and sends them to the sender. The sender responds with its encrypted indicators and the decrypted receiver's indicators. 
Afterwards, the receiver computes and stores the intersection result. 
For the OT-based method, a \textit{sender} and a \textit{receiver} are involved. The sender generates $k$ oblivious pseudo-random function (OPRF) seeds. The receiver applies a distinct pseudo-random function to each element it possesses, resulting in a specific mapped set. Simultaneously, the sender employs each pseudo-random function for every element in its possession, thereby generating its individual mapped set. Subsequently, the sender transmits its mapped set to the receiver, who conducts comparisons between the two sets for each element to derive the ultimate result.

\stitle{Multi-party PSI.} Based on the basic two-party PSI protocol, Tree-MPSI works as follows:

{\bf Step 1: Client request.} 
Clients request $R_{c}$ to the aggregation server to initiate data alignment and check the PSI process status (\ding{192}).

{\bf Step 2: Scheduling.}
We first identify the active clients, denoted as set $\mathcal{U}$. 
These clients are paired sequentially according to their request order to the server. For each pair, the earlier requester is assigned the role of sender while the later requester is receiver for the two-party PSI protocol (TPSI) (\ding{193}). 

{\bf Step 3: Server status generation.} 
Upon receiving client requests and completing scheduling, the aggregation server generates and sends a status message $R_{s}^{c}$ to each client to  notify its TPSI partner's communication address (\ding{194}).

{\bf Step 4: TPSI Computation.} 
After obtaining the communication address of its TPSI partner, the client engages in interaction with it. If the address is not available, the client waits until the entire Tree-MPSI process is complete to receive the final result.
(\ding{195}).

{\bf Step 5: Result allocation.} 
In the final round of Tree-MPSI, the client holding the final result $[N_{align}]$, which is an ordered list, encrypts the result using homomorphic encryption (HE) with the public key allocated by the key server:
$
Enc([N_{align}]) = HE.Enc([N_{align}], pk).
$
The client sends $Enc([N_{align}])$ to the aggregation server (\ding{196}), and the aggregation server forwards the encrypted result to all clients (\ding{197}), where the result can be decrypted using their private keys as: 
$[N_{align}] = HE.Dec(Enc([N_{align}]), sk)$.

\stitle{Comparing with alternatives.}
In contrast to the Path-MPSI requiring $O(m)$ communication rounds, Tree-MPSI achieves parallelism by scheduling and pairing clients.
Thus, it allows concurrent two-party PSI (TPSI) and reduces communication rounds to $O(\log_m)$.
During Tree-MPSI execution, clients merely communicate whether they have stored the TPSI result from the previous round and the current TPSI result's length they possess. Furthermore, the final results transmitted through the aggregation server are encrypted using homomorphic encryption (HE), ensuring that the aggregation server lacks the private key. 
Consequently, sensitive client data remains secure, and the RSA algorithm ensures privacy during TPSI computations.

\stitle{Scheduling optimization.} 
In two-party PSI, if RSA blind signature is chosen, the receiver encrypts and transmits the indicator information twice, whereas the sender does only once.
To reduce communication overhead, we pair a client that has a large dataset ($ResLen$ indicated in the requests) with a client that has a small dataset. The idea is to reduce the communication volume by assigning the client with fewer samples as receiver. If the protocol is OT-based,
since the sender needs to transmit a large amount of data, the client with a larger dataset is designated as the receiver.
In particular, the scheduling works as follows:

    \begin{itemize}
        \item \textit{Sorting.} We sort the active clients  in $\mathcal{U}$ by $ResLen$ in ascending order, yielding a sorted list
        $L_{unscheduled} = AsSort(\mathcal{U}) = [c_{1}, c_{2}, ...c_{|\mathcal{U}|}]$,
        where function $AsSort(\cdot)$  sorts the input array of clients in ascending order.
        
        \item \textit{Pairing.}
        We generate pairs of active clients from $L_{unscheduled}$. Specifically, we adopt a greedy approach, which pairs client $c_{k}$ with client $c_{k + \lceil\frac{|\mathcal{U}|}{2}\rceil}$ for $k=1, 2, ..., \lfloor{\frac{|\mathcal{U}|}{2}}\rfloor$. 
        When $|\mathcal{U}|$ is odd, client $\lceil\frac{|\mathcal{U}|}{2}\rceil$ is paired with itself.

        \item \textit{Selecting of two-party PSI result receiver.} 
        For RSA-based two-party protocol, the client with smaller amount of data is assigned as receiver. 
        If using an OT-based protocol, the client with more data is the receiver, cutting down two-party PSI communication overhead.
    \end{itemize}

\stitle{Performance analysis.} The scheduling optimization involves sorting the active clients with $O(|\mathcal{U}|\log_{|\mathcal{U}|})$ time and looping over the clients for pairing with O({$|\mathcal{U}|$}) time. Thus, the overall time complexity of the scheduling optimization is $O(|\mathcal{U}|\log_{|\mathcal{U}|})$.
As for the communication cost, we use $B$ to denote the number of samples in the larger dataset and $S$ for the number of samples held by the other party. 
If clients are paired based on their request order, the worst-case communication cost of two-party PSI is $O(2|B|+|S|)$. 
With our proposed optimization, the communication cost is $O(2|S|+|B|)$, which yields a reduction of $O(|B|-|S|)$. As the number of clients $|\mathcal{U}|$ is usually far fewer than the number of samples, the scheduling cost is much smaller than the saving of communication cost. 

\subsection{Cluster-Coreset for Coreset Construction}

 \begin{figure*}[!t]
 \centering
    \includegraphics[width=0.68\textwidth]{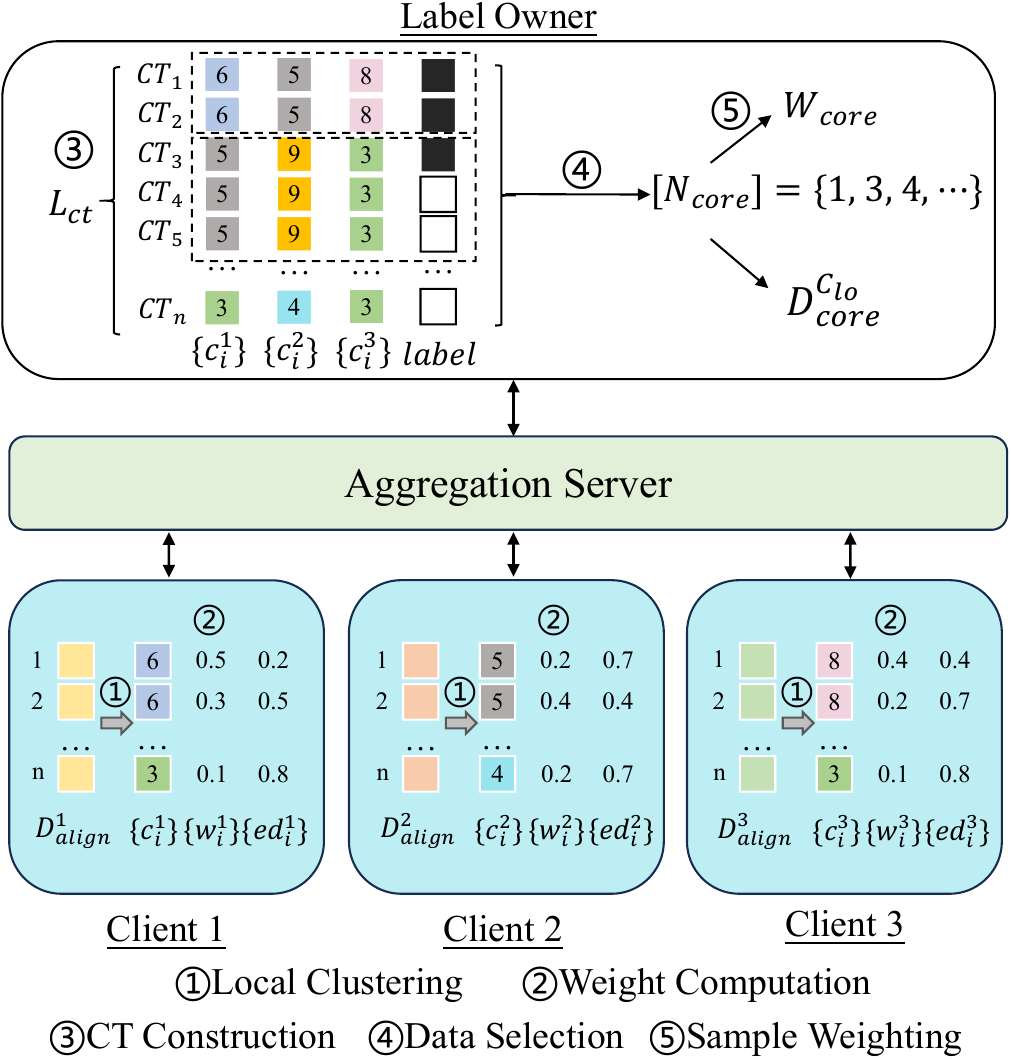}
    \caption{An illustration of Cluster-Coreset for coreset construction.}
    \label{coreset}
\end{figure*}
% \vspace{-1cm}

After executing Tree-MPSI, we denote the aligned samples as 
$D_{align}=\{(x_{i}\in\mathbb{R}^{d}, y_{i}\in\mathbb{N}): i \in [N_{align}]\}$. 
The features held by client $m$ can be denoted as 
$D_{align}^{m}=\{x_{i}^{m}\in\mathbb{R}^{d_{m}}: m \in[M], i \in [N_{align}]\}$, and the labels held by the label owner as 
$D_{align}^{label}=\{y_{i}\in\mathbb{N}: i \in [N_{align}]\}$. 
Following, we use the indicators $\{i, i \in [N_{align}]\}$ to refer to the aligned data samples. 

Our proposed coreset construction method takes a clustering strategy ---
first cluster samples on each client and then merge local clustering results.
Cluster-Coreset only communicates cluster indices and therefore can well protect data privacy.
Besides, Cluster-Coreset does not rely on the downstream tasks, and hence achieves better generalization compared to V-coreset.
The procedure of Cluster-Coreset contains the following 5 steps.

% {\bf Step 1: XXX}

% \begin{itemize}
{\bf Step 1: Local clustering.}
We perform separate clustering for the aligned samples on each client, which often represents a specific domain. 
Specifically, each client clusters its local features individually using the $K$-Means algorithm, generating $c$ clusters (\ding{192}).

{\bf Step 2: Weight computation.} 
Post local clustering, each client's samples are weighted according to their distance to the cluster centers, with nearer samples receiving higher weights (\ding{193}).
Let $S_m^{c}$ be the set of samples that belong to cluster $c$ on client $m$, the weight $w_{i}^{m}$ of each sample $i \in S_m^{c}$ is computed as:
$$
w_{i}^{m}= \frac{1}{|S_m^c|} \times {pos}(ed_{i}^{m}, {DeSort}(\{ed_{j}^{m}\}, j \in {S_m^c})),
$$
where function ${DeSort}(\cdot)$ sorts the samples in descending order based on their Euclidean distances $ed_{i}^{m}$ to the cluster centroids,  
${pos}(\cdot, \cdot)$ returns the position of sample $i$ in the sorted array. 
The weights reflect the significance of the samples, that is, those closer to the centroids are more representative.
    
{\bf Step 3: Cluster tuple (CT) construction.}
Each client $m$ sends HE-encrypted messages to the label owner via the aggregation server. For each sample $i$, the message contains the local weight $w_{i}^{m}$, cluster index $c_{i}^{m}$, and the Euclidean distance to the cluster centroids $ed_{i}^{m}$. 
The aggregation server concatenates the messages for the same sample before sending them to the label owner. This ensures that the label owner cannot infer the sources of the messages. The label owner uses the messages to construct the CTs. In particular, for each data sample $i$ in $[N_{align}]$, we gather its cluster indices from all clients as:
$$
CT_{i} = (c_{i}^{1}, c_{i}^{2}, \ldots, c_{i}^{m}).
$$
Thus, the CTs list for all $D_{align}$ samples is $L_{ct} = {CT_{i}: i \in [N_{align}]}$ (\ding{194}).

{\bf Step 4: Data selection.}
After constructing CTs, the next question becomes: \textit{how to choose a small group of samples as the coreset?}
Intuitively, if the CTs of two samples are identical, it indicates that these samples have been classified in the same cluster across all clients. 
%In other words, irrespective of the client, these samples can be considered similar. 
Therefore, they are considered as similar, and we select only one sample to be included in the final coreset.
By utilizing $L_{ct}$, the label owner identifies the most informative and representative samples. 
We calculate the set of distinct CT values in $L_{ct}$ as $T_{ct} = \{t_1, t_2,\cdots, t_k\}$, where $k$ is the number of distinct CT values. 
We use $S_{ct}^{j}$ to denote the samples with value $t_j$, which is expressed as:
$
S_{ct}^{j}=\{ i \in [N_{align}]: CT_{i}=t_{j} \}.
$
We then split $S_{ct}^{j}$ into subsets based on their labels:
$
S_{ct}^{j, l} = \{ i \in  S_{ct}^{j}: y_{i}=l \}.
$
%After splitting the sets and obtaining $S_{ct}^{j,l}$, 
The label owner needs to select one sample from $S_{ct}^{j,l}$. 
In particular, the sample with the minimal aggregated distance to its cluster centers (i.e. $\sum_{i}^{m} ed_i^m$) is chosen. The chosen sample's indicator can be expressed as:
$$
i_{j, l}^{*} = argmin_{i \in S_{ct}^{j, l}} \sum_{m=1}^{M} ed_{i}^{m}.
$$

In this way, the label owner obtains the sample indicators to be included in the coreset, i.e., $[N_{core}] = \{i_{1, 1}^{*}, i_{1, 2}^{*}, \ldots, i_{k, L}^{*}\}$, where $L$ is the number of distinct classes held by the label owner (\ding{195}). The selected indicators are then encrypted with HE and sent to all clients via the aggregation server. 
Each client utilizes the decrypted indicators to select the corresponding local data samples and collaboratively construct the coreset $D_{core} = \{(x_{i} \in \mathbb{R}^{d}, y_{i} \in \mathbb{N}): i \in [N_{core}]\}$.

{\bf Step 5: Sample weighting.}
Recall Step 3, we assign a weight for each sample to reflect its importance/confidence.
We propose to weight the coreset samples during model training (\ding{196}). 
In particular, the weight of each coreset sample is obtained by summing its local weights: 
$
w_{i} = \sum_{m=1}^{M} w_{i}^{m}, i \in [N_{core}].
$
Intuitively, the weighting strategy adjusts the importance of the samples according to their proximity to the cluster centroids and considers samples close to the centroids as more important. 
With the weights, the loss function for classification and regression tasks can be expressed as
\begin{equation}
L(D_{core}, W_{core}, \theta):=\sum_{i \in[N_{core}]} w_{i} \cdot L(x_{i}, \theta).
\end{equation}

\noindent \textbf{Privacy analysis.}
In the process of constructing the coreset, the transmitted data, including the weights, cluster categories, distances, and the selected data indicators, are all encrypted using homomorphic encryption (HE).
As the aggregation server does not know the decryption key, the entire process is secure against a curious server.
In contrast, V-coreset sends the original labels to the server for model
training, potentially resulting in label leakage.

\section{Experimental Evaluation}

% \vspace{-0.8cm}

\begin{table} 
\caption{Statistics of the experiment datasets.}
\centering
\scalebox{1}{
    \begin{tabular}{ccccccc}
        \toprule  
            \textbf{Dataset} & \textbf{BA} & \textbf{MU} & \textbf{RI} & \textbf{HI} &  \textbf{BP} & \textbf{YP}  \\ 
        \midrule 
            \# instances & 10K & 8K & 18K &  100K &13K  & 510K \\
            \# features & 11 & 22 & 11 & 32 & 11 & 90 \\ 
            \# classes & 2 & 2 & 2 & 2 & 4 & / \\ 
            % \# partitions & 3 & 3 & 3 & 3 & 3 & 3\\
        \bottomrule 
    \end{tabular}
}
    % \caption{Dataset descriptions}
    \label{table:dataset}
\end{table}

% \vspace{-1cm}

\begin{table*}[ht]
\caption{Framework comparison w.r.t model accuracy and runtime (best in bold).}

\centering
\renewcommand{\arraystretch}{1.3}
\scalebox{0.99}{
\resizebox{\linewidth}{!}{
\scriptsize

\begin{tabular}{c|c|cc|cc|ccc|ccc|c|c}
\hline
\multirow{2}{*}{}                                                      & \multirow{2}{*}{\textbf{Method}} & \multicolumn{2}{c|}{\textbf{BA}}           & \multicolumn{2}{c|}{\textbf{MU}}    & \multicolumn{3}{c|}{\textbf{RI}}                        & \multicolumn{3}{c|}{\textbf{HI}}                             & \textbf{BP}   & \textbf{YP}           \\ \cline{3-14} 
                                                                       &                         & LR               & MLP              & LR               & MLP            & LR             & MLP            & KNN            & LR               & MLP              & KNN              & MLP             &
                                                                       LinearReg \\ \hline
\multirow{2}{*}{\begin{tabular}[c]{@{}c@{}}Acc (\%) /\\ MSE\end{tabular}}                                               & ALL                  & 80.5\%          & 85.3\%          & 95.3\%          & 95.5\%        & 100\%          & 100\%          & 99.9\%        & 99.0\%          & 99.3\%          & \textbf{88.9\%} & 66.0\%     & \textbf{90.59}    \\
                                                                       & CSS                   & \textbf{81.1\%} & \textbf{85.7\%} & \textbf{97.2\%} & \textbf{100\%} & \textbf{100\%} & \textbf{100\%} & \textbf{100\%} & \textbf{99.2\%} & \textbf{99.9\%} & 88.8\%           & \textbf{66.2\%} &
                                                                       90.63
                                                                       \\

\hline
\multirow{4}{*}{Time (s)}                                               & \starall                  & 284             & 346             & 203             & 253           & 220           & 322           & 128           & 1730            & 2085            & 1730            & 523     & 60000       \\
& \treeall                  & 250             & 320             & 187             & 230           & 206          & 290          & 120           & 1600            & 1985            & 1664            & 484    & 57000       \\
& \starcss                  & 187             & 203             & 140            & 154           & 98           & 130          & 103         & 1350            & 1408            & 1276            & 365     & 52400       \\
                                                                       & \sys                   & \textbf{124}    & \textbf{159}    & \textbf{81}     & \textbf{123}  & \textbf{75}   & \textbf{103}  & \textbf{100}  & \textbf{1275}   & \textbf{1212}   & \textbf{1198}   & \textbf{267}   &
                                                                       \textbf{48000}
                                                                       \\

\hline
\multirow{2}{*}{\begin{tabular}[c]{@{}c@{}} Train \\ Data\end{tabular}} & ALL                 & 7000             & 7000             & 5686             & 5686           & 12729          & 12729          & 12729          & 70000            & 70000            & 70000            & 9375     & 463715       \\
                                                                       & CSS                   & \textbf{4237}    & \textbf{4152}    & \textbf{1636}    & \textbf{2268}  & \textbf{203}   & \textbf{285}   & \textbf{285}   & \textbf{33100}   & \textbf{31036}   & \textbf{32924}   & \textbf{6413}  &
                                                                       \textbf{71059}
                                                                       
                                                                       \\

 \hline
\end{tabular}
}}
% \caption{Evaluation of \sys under different datasets}
\label{table:performance}
\end{table*}

\begin{figure*}[!]
    \centering
    \includegraphics[width=1.\textwidth]{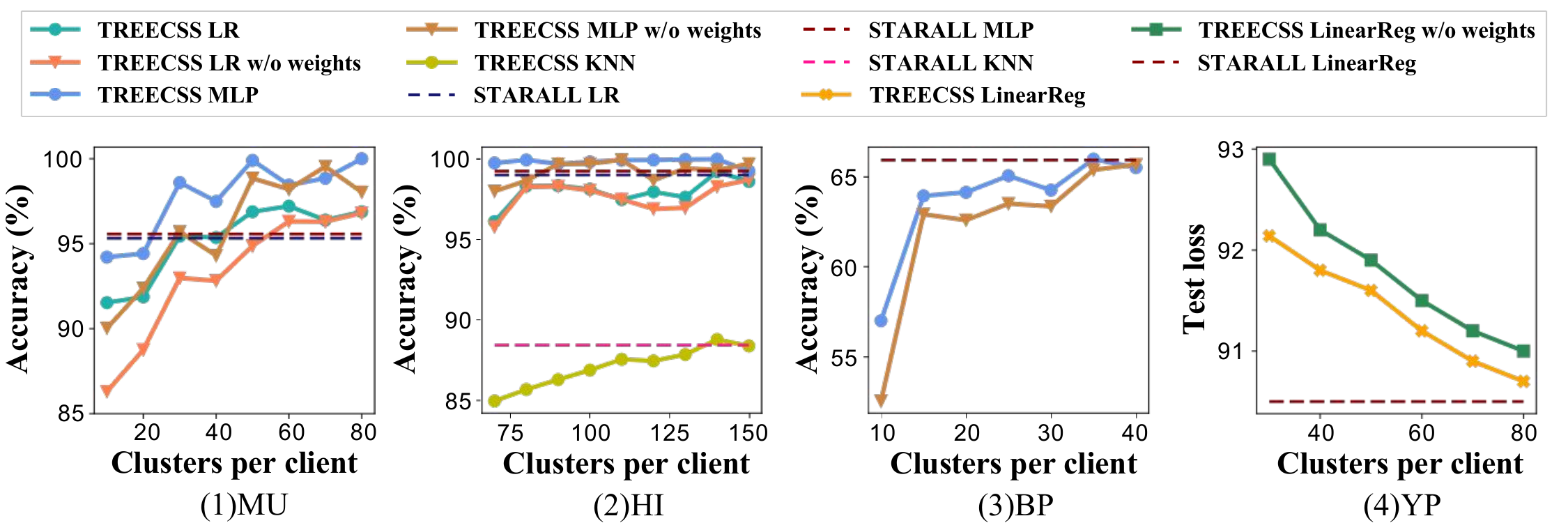}
    \caption{Effect of cluster size and re-weighting on model quality.}
    \label{fig:ablation_acc}
\end{figure*}

We conduct extensive experiments on diverse datasets stored in Oceanbase~\cite{yang2023oceanbase,yang2022oceanbase} to validate the effectiveness of our proposed \sys. 
%Firstly, we evaluate \sys and the baselines over both classification and regression tasks.
%We then evaluate Tree-MPSI using a large number of clients and various data scales. 

\subsection{Experiment Settings}

\stitle{Datasets.}
Table \ref{table:dataset} presents an overview of the datasets evaluated in this work. Among them, Bank (BA) \cite{BankDataset}, Mushrooms (MU) \cite{MushroomsDataset}, Rice (RI) \cite{RiceDataset}, and Higgs (HI) \cite{HiggsDataset} are binary classification datasets.
Note that, we randomly sample 100000 data samples from the HI dataset.
BodyPerformance (BP) \cite{BodyPerformanceDataset} is a multi-class classification dataset, containing four distinct categories. 
YearPredictionMSD (YP)~\cite{misc_yearpredictionmsd_203} is a  regression dataset. 
%We employed this dataset in our experiments to further demonstrate the scalability of our proposed method in linear regression tasks.
For classification datasets, we divided the data into a training set (70\%) and a test set (30\%). As for the YP dataset, following the instructions provided by the author, we divided it into a training set of 463,715 data points and a test set of 51,630 data points.

\stitle{Models.} 
We choose {\it logistic regression} (LR), {\it multi-layer perceptron} (MLP) with one hidden layer, and {\it $k$-nearest neighbors} (KNN) as the downstream classification ML models.
For LR, we conduct experiments on BA, MU, RI, and HI. For MLP, we run on BA, MU, RI, HI, and BP. As for KNN, we choose RI and HI as the representative datasets.
For the regression task, we choose {\it linear regression} (LinearReg) model.

\stitle{Baselines.} We compare our proposed methods with the following baselines:
\begin{itemize}
    \item For evaluating the entire framework, we include a VFL process utilizing the star-structured MPSI protocol and training all available data via SplitNN, referred to as {\em \starall}. 
    {\em \treeall} means Tree-MPSI combined with all training data. 
    {\em \starcss} consists of Star-MPSI and Cluster-Coreset. 
    Our {\em \sys} includes Tree-MPSI, Cluster-Coreset construction, and the model training process. 
    
    \item For evaluating Tree-MPSI, we run different topologies including path and star for multi-party private set intersection (MPSI) in VFL. 
    %The path topology PSI involves conducting two-party PSI (TPSI)} operations between clients, with a step-by-step manner. 
    In path-topology PSI, each client conducts TPSI with the next client, and the process continues until all the clients have been involved.
    In a star topology, the central node runs TPSI separately with each of the remaining nodes.
    
    \item For evaluating coreset construction, we include V-coreset as a baseline for both classification and regression tasks. 
    
\end{itemize}
    
\stitle{Protocols.}
We launch a cluster with three clients and one label owner. 
The dataset is equally partitioned into three portions, and each portion is held by one client. Meanwhile, the label owner has all the labels of the dataset.
We choose Adam~\cite{kingma2014adam} as the optimization algorithm for both classification and regression tasks. 
We perform grid search to tune the learning rate within $\{1, 0.1, 0.01, 0.001\}$ across all datasets. We tune the batch size from 0.1\% to 1\% of the training data. 
We iteratively train the downstream model until loss change over 5 epochs is below the threshold of 1e-4, indicating convergence.
To ensure a fair comparison, we initialize basic TPSI in all baselines as \textit{RSA-based}, unless otherwise specified.

\stitle{Implementation.} 
We utilize Numpy and PyTorch for the model training. For communication, we choose the gRPC framework with proto3. Furthermore, we utilize the TenSEAL library for homomorphic encryption (HE) operations. All experiments are conducted on a cluster, with each machine equipped with 24 GB memory, 8 cores, and 10GBps bandwidth.

\subsection{End-to-end Performance}

% \vspace{-0.5cm}
\noindent \textbf{Model performance.}
We compare the baselines and \sys in terms of model quality. As Star-MPSI and Tree-MPSI do not impact the model's quality, we only compare ALL and CSS.
% Table \ref{table:performance} shows that CSS consistently achieves comparable or higher accuracy across various datasets on the LR and MLP models, with improvements of up to 2\% and 4.5\% respectively. 
% CSS also demonstrates competitive performance on KNN by leveraging the coreset for similarity computation.  
% On regression task, CSS obtains similar test MSE on the YP dataset compared to ALL.
Table 2 reveals that CSS attains similar or greater accuracy on diverse datasets using LR and MLP models, showing up to 2\% and 4.5\% enhancements, respectively. It also performs well with KNN through coreset-based similarity calculations and achieves comparable test MSE on the YP dataset in regression tasks as compared to ALL.

\noindent \textbf{Time consumption.}
In terms of time consumption, by reducing the training data size, \sys considerably speeds up the training process. 
For example, on RI dataset, \sys accelerates training by up to 2.93x compared with \starall. 
On average, \sys requires only about 54\% of the original training time for downstream models.

% \jiang{TODO: move the following to CSS comparison.}
% In addition, as showm in Table.~\ref{table:performance}, our Cluster-Coreset effectively reduces the amount of data involved in the final training, up to 98.4\% in RI dataset, thereby reducing the training burden.

\begin{figure*}[!t]
    \centering
    \includegraphics[width=1\textwidth]{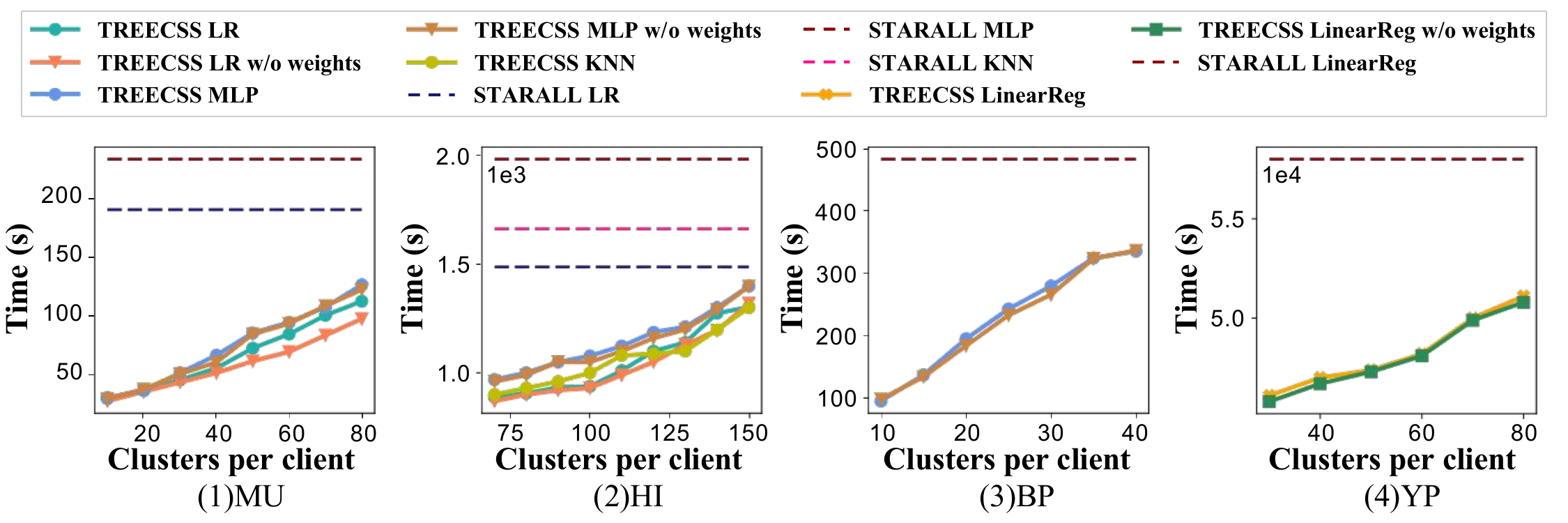}
    \caption{Effect of cluster size and re-weighting on runtime.}
    \label{fig:ablation_time}
\end{figure*}

\begin{figure*}[!t]
    \centering
    \includegraphics[width=1\textwidth]{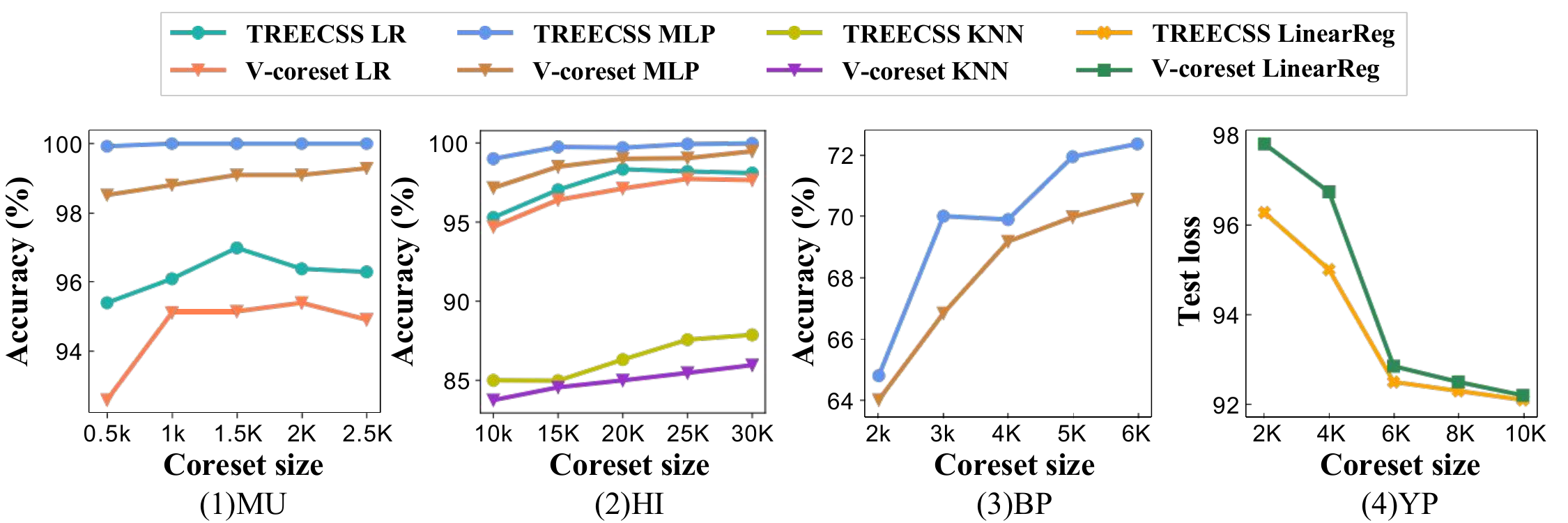}
    \caption{Comparing the model quality of V-coreset and \sys.}
    \label{fig:vs_vcoreset}
\end{figure*}

\subsection{Evaluation of Tree-MPSI and Cluster-Coreset}

\begin{figure}[!t]
    \centering
    \subfigure[RSA-based protocol]
    {
        % \begin{minipage}[t]{0.45\linewidth}
        \hspace{-0.6cm}\includegraphics[width=0.32\textwidth]{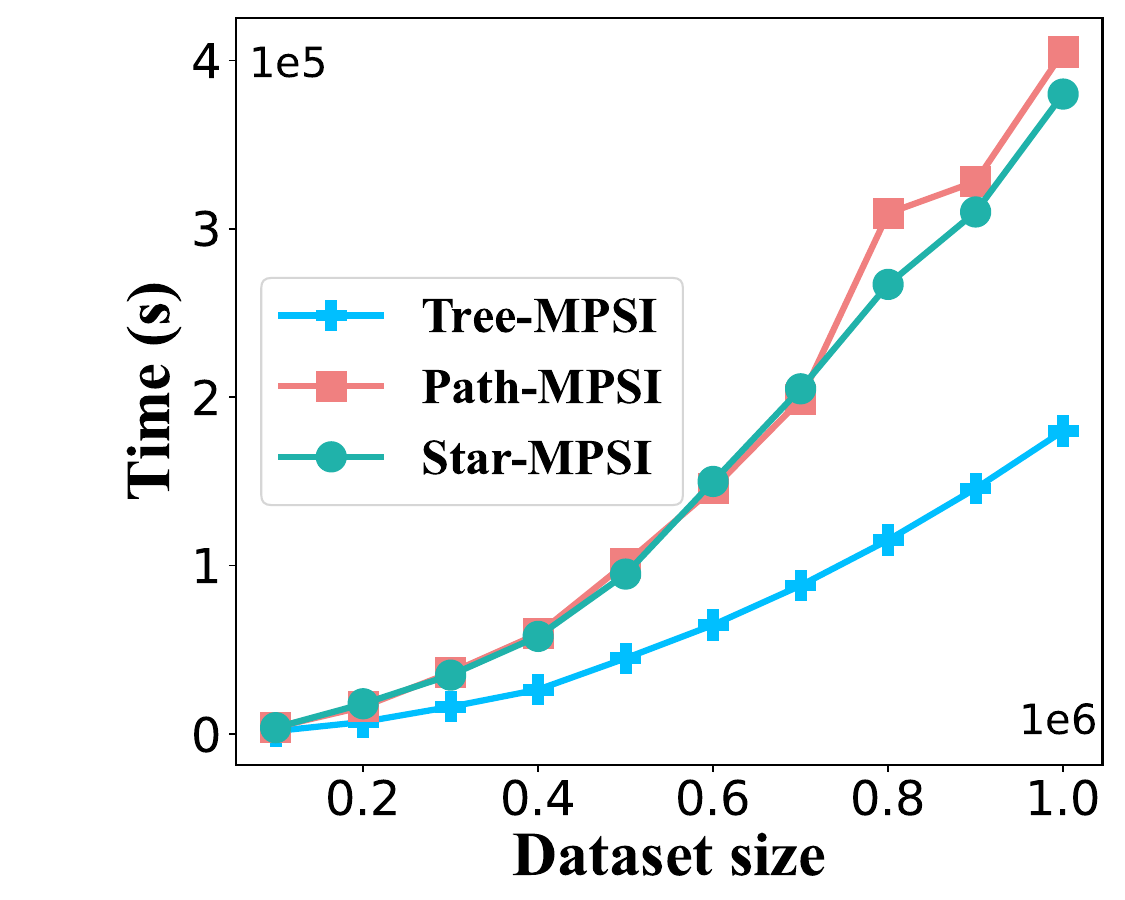}
        % \end{minipage}
    \label{fig:bt-mpsi_vs_baseline_rsa}
    }
    \subfigure[OT-based protocol]{
        % \begin{minipage}[t]{0.45\linewidth}
        \hspace{0cm}\includegraphics[width=0.32\textwidth]{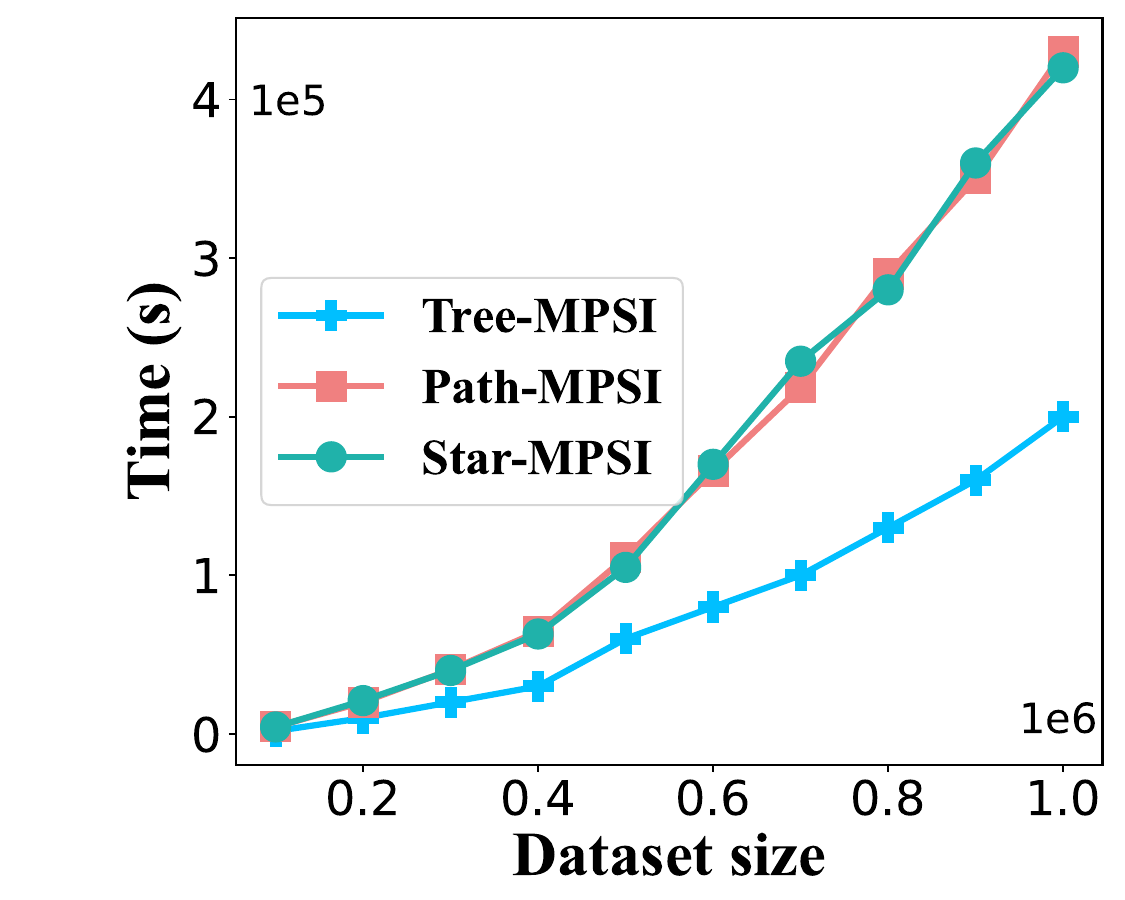}
        % \end{minipage}
    \label{fig:bt-mpsi_vs_baseline_ot}
    }
    \subfigure[Scheduling optimization]{
        % \begin{minipage}[t]{0.45\linewidth}
        \includegraphics[width=0.32\textwidth]{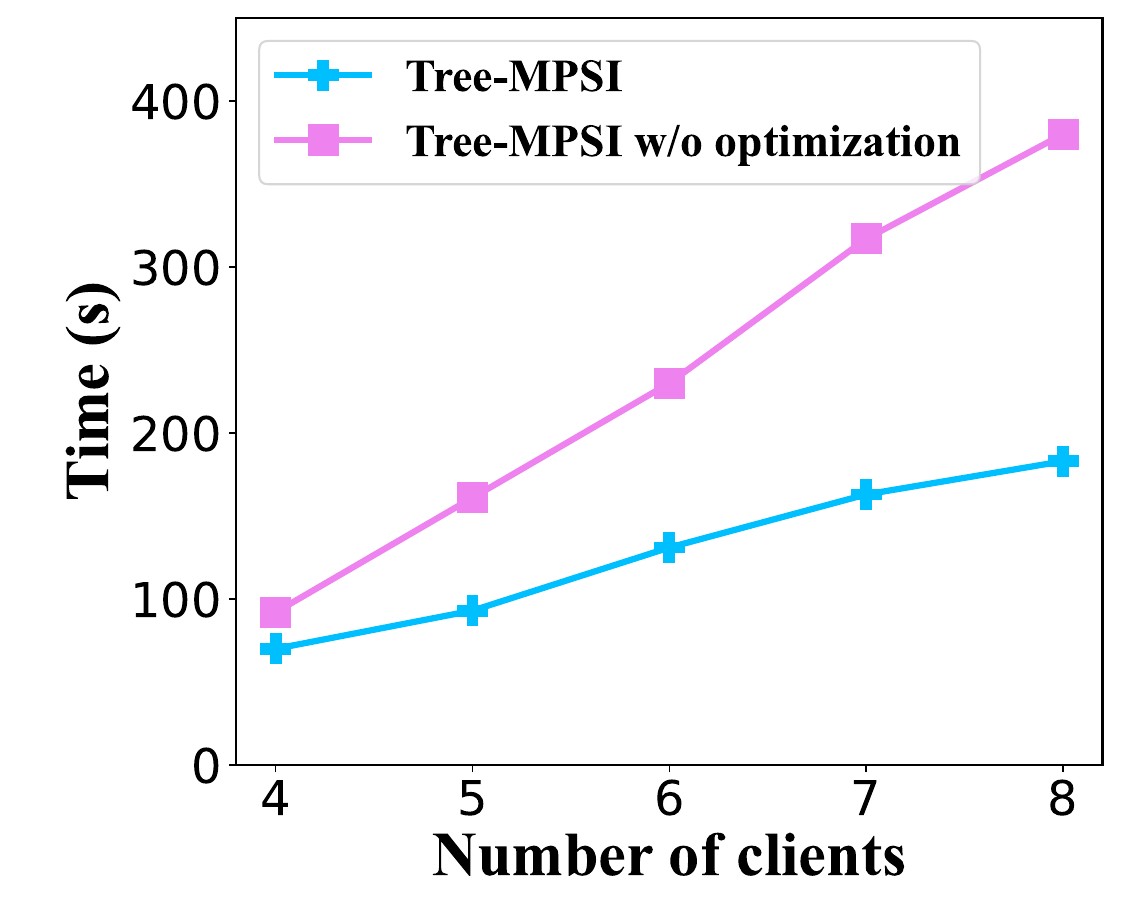}
        % \end{minipage}
    \label{fig:bt-mpsi_optimization}
    }
    
    \caption{Evaluation of Tree-MPSI. (a) and (b) use different TPSI protocols, and (c) sets the number of samples on client $i$ as 10000$\times i$.}

\end{figure}

We next evaluate two components in \sys: Tree-MPSI and Cluster-Coreset.

\stitle{Tree-MPSI.}
We generate a synthetic dataset that only has data sample indicators for each client. 
The content within these datasets overlaps by 70\%, and each client's indicators are randomly shuffled. 
Initially, we compare Tree-MPSI and the baselines (Path-MPSI, Star-MPSI) with 10 clients across varying dataset sizes per client, as illustrated in Fig~\ref{fig:bt-mpsi_vs_baseline_rsa} (RSA-based) and \ref{fig:bt-mpsi_vs_baseline_ot} (OT-based). 
The results show an average 2.25$\times$ speedup with Tree-MPSI, and the improvement becomes more significant as the dataset size increases.
We further validate our scheduling optimization in Fig. \ref{fig:bt-mpsi_optimization}. 
We let client $i$ hold 10,000$\times i$ data points (e.g., the rank-2 client has 20,000 data points). Without our optimization, the clients are paired sequentially according to their request order. The results demonstrate the efficiency of our volume-aware scheduling method in pairing two-party PSI, and its effectiveness becomes more pronounced as the number of clients increases.

\stitle{Cluster-Coreset.}
% In VFL, it is possible that the label-holding parties may not possess data features. However, the VRLR framework assumes the presence of labels in one of the data feature-holding parties when constructing the coreset. Therefore, after removing the labels from the coreset construction process of the VRLR framework, we proceed to compare the model performance of \sys against it.
We compare V-coreset and Cluster-Coreset w.r.t. model quality.
We do not report time cost since V-coreset has not implemented their method in a distributed manner.

As shown in Fig. \ref{fig:vs_vcoreset}, 
on both classification and regression tasks, under the same coreset size, \sys demonstrates better testing performance than V-coreset.
Additionally, we report the performance of Cluster-Coreset on reducing the dataset volume, 
up to 98.4\% in RI dataset, thereby effectively reducing the training burden.

\subsection{Ablation and Sensitivity Study}

\stitle{Clusters per client.}
We assess \sys by varying the number of clusters per client on 4 representitive datasets: MU, HI, BP and YP.
% As depicted in Fig. \ref{fig:ablation_acc}, 
% larger cluster size can improve test accuracy and reduce test loss, attributed to the inclusion of more data samples in the coreset.
% Meanwhile, increasing the cluster size leads to increased time consumption due to a larger coreset, as shown in Fig. \ref{fig:ablation_time}.
Fig. \ref{fig:ablation_acc} shows that larger cluster sizes enhance test accuracy and lower test loss by including more samples in the coreset, but as Fig. \ref{fig:ablation_time} indicates, this also raises time consumption due to a larger coreset's size.

\stitle{Effect of reweighting.}
We next evaluate the effect of reweighting. 
As shown in Fig.~\ref{fig:ablation_acc}
and~\ref{fig:ablation_time}, the introduction of weights significantly enhances model test performance, particularly with fewer clusters. 
The reweighting mechanism slightly prolongs training time compared to training without weights.

\iffalse
\stitle{Clusters per client and w/o weights.}
We choose 4 representitive datasets to conduct the ablation study: Mush, Higgs, BP and YP.
After validating \sys's efficiency in accelerating VFL training while maintaining model accuracy, we assess \sys alongside two variations: clusters per client and the absence of weights (w/o weights). Additionally, the introduction of weights significantly impacts model test performance, particularly with fewer clusters. However, using weights during model training slightly prolongs training time compared to training without weights.
\fi

\section{Conclusions}

In this paper, we present \sys, an efficient end-to-end VFL framework. This approach aims to address the limitations of prior methods, which lack efficient data alignment, privacy protection and the support of various task. 
To enhance end-to-end VFL, we a Tree-MPSI protocol for the alignment stage and a clustering coreset construction method for the training stage, coupled with a reweighting strategy.
Extensive experiments show that our proposed framework significantly outperforms the baselines.

\section*{Acknowledgment}
% This work was sponsored by Key R\&D Program of Hubei Province (No. 2023BAB \\ 077, No. 2023BAB170), the Fundamental Research Funds for the Central Universities (No. 2042023kf0219), and CCF- AFSG Research Fund. 
% This work was supported by Ant Group through CCF-Ant Research Fund (CCF-AFSG RF20220001).
This work was sponsored by Key R\&D Program of Hubei Province (No. 2023BAB
077, No. 2023BAB170), and the Fundamental Research Funds for the Central Universities (No. 2042023kf0219). This work was supported by Ant Group through CCF-Ant Research Fund (CCF-AFSG RF20220001).

\bibliographystyle{splncs04}
\bibliography{referances}

\end{document}